\documentclass[runningheads]{llncs}
\usepackage{times}
\usepackage{epsfig}
\usepackage{graphicx}
\usepackage{amsmath}
\usepackage{amssymb}
\usepackage{color}
\usepackage{siunitx}
\usepackage{subcaption}

\usepackage[pagebackref=true,breaklinks=true,colorlinks,bookmarks=false]{hyperref}

\begin{document}

\title{CNN-Based Semantic Change Detection in Satellite Imagery\thanks{A. Gupta is funded by the School of Electrical and Electronic Engineering, The University of Manchester and the ACM SIGHPC/Intel Computational and Data Science Fellowship. E. Welburn is funded by the EPSRC HOME Offshore project grant EP/P009743/1}}

%
\author{Ananya Gupta\orcidID{0000-0001-6743-1479} \and
Elisabeth Welburn\orcidID{0000-0003-3236-6755}  \and
Simon Watson\orcidID{0000-0001-9783-0147} \and
Hujun Yin\orcidID{0000-0002-9198-5401}}
\authorrunning{A. Gupta et al.}
%
\institute{The University of Manchester, Manchester, UK
\email{\{ananya.gupta, elisabeth.welburn, simon.watson, hujun.yin\}@manchester.ac.uk}}
%
\maketitle      


\begin{abstract}
   Timely disaster risk management requires accurate road maps and prompt damage assessment. Currently, this is done by volunteers manually marking satellite imagery of affected areas but this process is slow and often error-prone. Segmentation algorithms can be applied to satellite images to detect road networks. However, existing methods are unsuitable for disaster-struck areas as they make assumptions about the road network topology which may no longer be valid in these scenarios. Herein, we propose a CNN-based framework for identifying accessible roads in post-disaster imagery by detecting changes from pre-disaster imagery. Graph theory is combined with the CNN output for detecting semantic changes in road networks with OpenStreetMap data. Our results are validated with data of a tsunami-affected region in Palu, Indonesia acquired from DigitalGlobe.
\end{abstract}

\section{INTRODUCTION}

\begin{figure}[]
\centering
\begin{subfigure}[]{0.45\textwidth}
\centering
  \includegraphics[width=\textwidth]{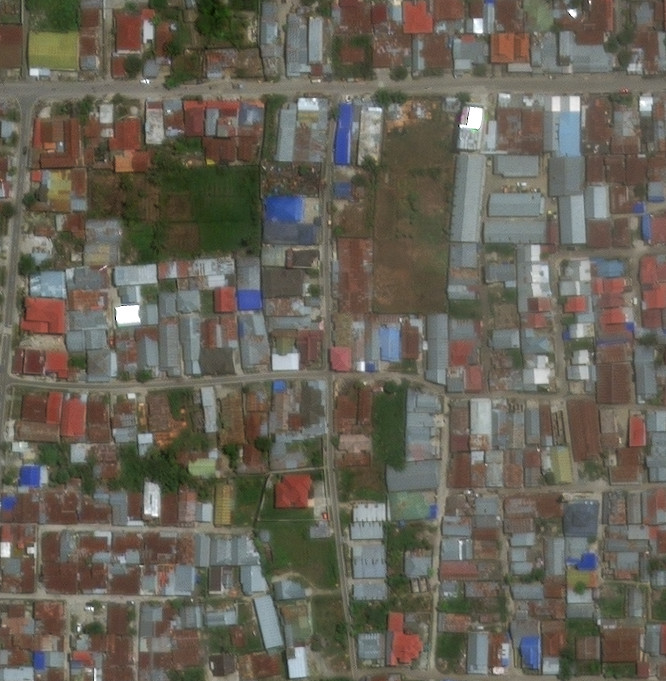}
\end{subfigure} \hfill
\begin{subfigure}[]{0.45\textwidth}
  \centering
  \includegraphics[width=\textwidth]{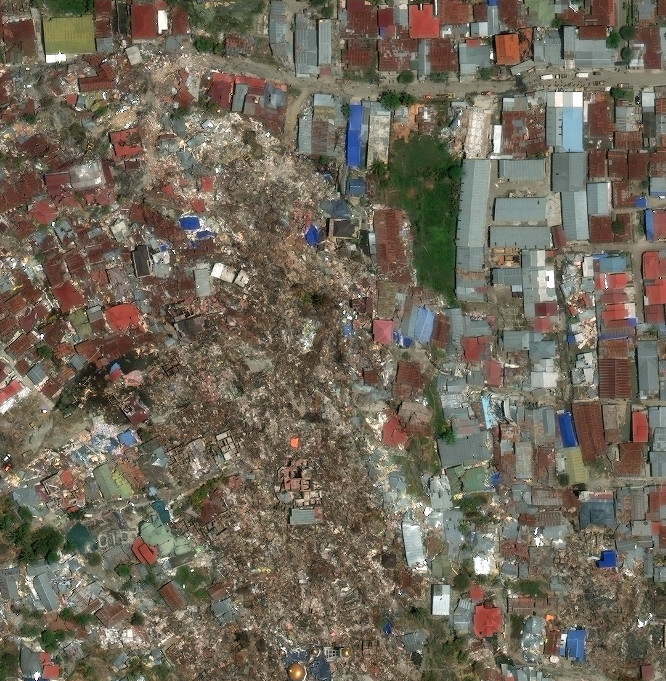}
\end{subfigure}
\caption{Extracted images from satellite imagery of Palu, Indonesia showing the devastation due to the tsunami and earthquake in September, 2018~\cite{DIGI}. \textit{Left}: Before the tsunami. \textit{Right}: The day after the tsunami.}
\label{fig:dev}
\end{figure}

Hundreds of natural disasters strike every year across the globe\footnote{https://ourworldindata.org/natural-disasters}. Timely damage assessment and mapping of disaster-struck areas are extremely important to disaster relief efforts. They are especially important in developing countries where the affected areas may not have been mapped. Furthermore, routes into affected areas can be blocked due to the effects of the disaster, rendering pre-existing maps ineffectual. An example of a disaster-struck area is shown in Fig. \ref{fig:dev}.

Volunteer initiatives around the world make use of publicly available satellite imagery to map out such areas following natural disasters to help provide prompt assistance~\cite{Boccardo2015}. However, due to inconsistency across different initiatives and inexperience of the volunteers, this process is often error-prone and time-consuming~\cite{Poiani2016}.





There is increasing demand for automating the process of road extraction from satellite imagery since up-to-date road maps are important for location and navigation services~\cite{Miller2014}. The current research in road identification treats it as a semantic segmentation task where satellite images are used to predict the probability of a pixel being a road~\cite{Mattyus2017}. However, due to variations, shadows and occlusions caused by buildings and trees, a number of road segments are often misidentified by the segmentation algorithms. Heuristics based methods are often used to help alleviate this problem by reasoning about missing connections between broken roads. However, these assumptions do not work well for post-disaster scenarios as there are broken connections due to blockages and damages caused by the disaster.

Comparing pixel values of satellite imagery from before and after a disaster is a potentially useful approach for identifying the effect of the disaster. However, comparing pixels directly is implausible due to various effects such as illuminations and seasons, which can cause significant changes in image statistics, shadows and changes in vegetation. The use of Convolutional Neural Networks (CNNs) has been proposed for damage assessment in buildings to order to cope with these challenges~\cite{Amit2017,Fujita2017,Rudner2019}. However, these approaches require a large amount of manually annotated training data for each location, which is expensive, time-consuming and unscalable.


Hence, instead of comparing pixel values or requiring manually annotated data we propose to use data from OpenStreetMap (OSM)~\cite{Contributors2017} to train a CNN to identify semantic features such as roads in satellite imagery. This is used with pre-disaster and post-disaster imagery to help (a) identify the changes due to the disaster and identify the impacted areas; and (b) map out the road networks in the post-disaster landscape to aid disaster relief efforts. We further combine the OSM data with graph theory to obtain a more robust estimate of road networks. The framework allows for assessing the level of damage so as to identify high impact areas in a timely manner. A costing function is used to express the usability of affected roads for accessibility.







\section{RELATED WORK}

\subsection{Road Segmentation}
There are several existing approaches for extracting road maps from satellite imagery. A number of these approaches are based on probabilistic models. Geometric probabilistic models have been developed for road image generation followed by MAP estimation over image windows for road network identification~\cite{Barzohar1996}. Wenger \textit{et al.}~\cite{Wegner2015} proposed a probabilistic network structure to minimise a high order Conditional Random Field model to determine road connectivity. Another approach is based on manually identifying road points to define a road segment followed by matching further connected segments using a Kalman Filter~\cite{Vosselman2011}. Heuristic methods based on radiometric, geometric, and topological characteristics have been used to define road models, which are further refined based on contextual knowledge about objects such as buildings~\cite{Hinz2003} .

The drawback with both the heuristics and probabilistic approaches is that they work under an inherent assumption that road networks are connected and patches of roads do not exist in isolation. However, in post-disaster scenarios, this assumption is often invalid. Major points that we are interested in are actually the paths that are missing and/or broken connections left in the aftermath of the disaster. 

More recently, CNN-based methods have been used to segment road pixels from non-road pixels in satellite imagery. Some methods have approached this as a segmentation problem and reported pixel based metrics~\cite{Aich2018,Demir2018,Mnih2010}. However, since this approach does not take overall network topology into account, small gaps in the resultant network are not penalised, though they cause lengthy detours in practice. Another approach extracted topological networks from the segmented output and used smart heuristics in post processing to connect missing paths and remove small stubs which were seen as noise~\cite{Mattyus2017,VanEtten2018}. However, similar to the heuristic approach, it is not suitable for disaster-struck areas because the assumption of roads being connected is no longer valid.

\subsection{Disaster Analysis}
Satellite imagery is becoming an increasingly popular resource for disaster response management~\cite{Voigt2007}. Recent research has mostly focused on identifying buildings affected by floods and hurricanes. For instance, a CNN-based fusion of multi-resolution, multi-temporal and multi-sensor images was used to extract spatial and temporal characteristics for finding flooded buildings~\cite{Rudner2019}. CNNs have also been used to classify the probability of washed-away buildings by using clips of pre-disaster and post-disaster imagery~\cite{Amit2017,Fujita2017}.

Automated road identification in post-disaster scenarios is a nascent topic with concurrent research~\cite{Gupta2019}. Some work is based on identifying road obstacles such as fallen trees and standing water using vehicle trajectory data\cite{Chen2018}. Estimating road registration errors following earthquakes using post-event images has also been studied~\cite{Liu2019}. However, it could only correct for ground shifts due to earthquakes but could not address the problem of missing roads. A crowd-sourced pedestrian map builder was proposed~\cite{Bhattacharjee2019} but it would require people walking around in potentially inaccessible disaster-struck areas.

Similar to our work, Doshi \textit{et al.} \cite{Doshi2018} proposed a framework for change detection using satellite images in conjunction with CNNs. They identified buildings and roads in images before and after a disaster and used the per-pixel differences to quantify the disaster impact. By contrast, we focus on identifying road networks and correlating the changes with data obtained from OSM. We further propose a cost-based routing approach to take the affected areas into account when identifying possible routes for first responses.

\begin{figure*}[h!]
    \centering
    \includegraphics[width=\textwidth]{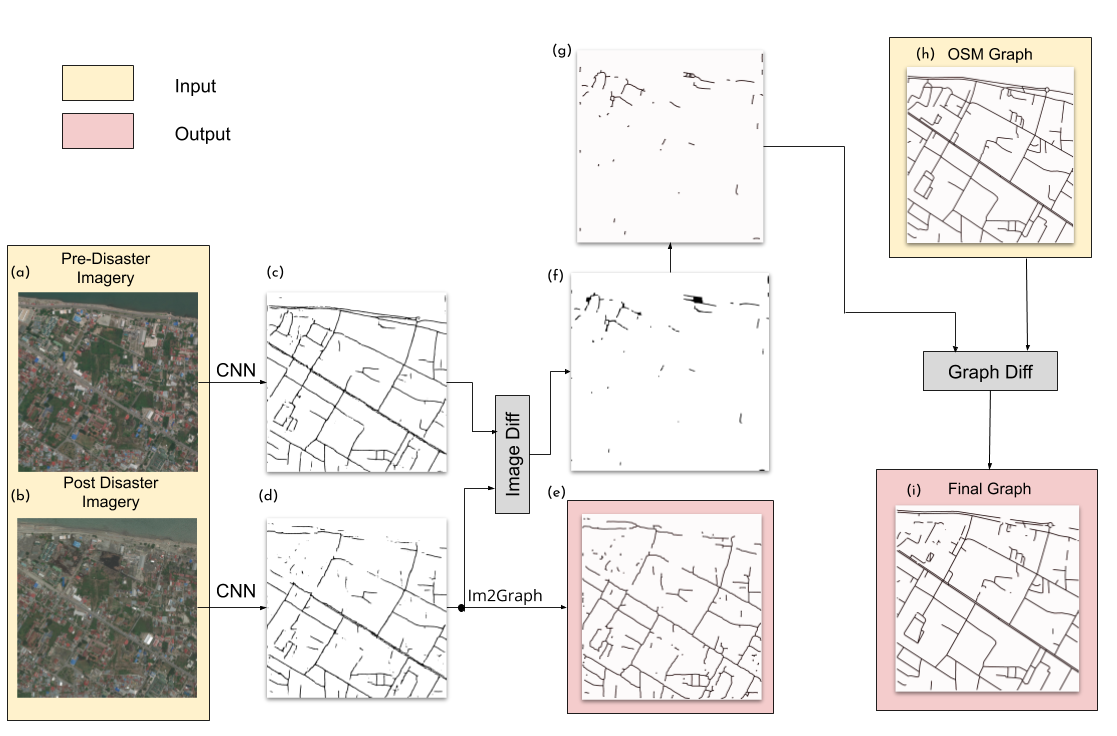}
    \caption{Block diagram of proposed methodology: pre-disaster satellite image (a) and post-disaster satellite image (b) are converted to road masks, (c) and (d), respectively using a CNN-based segmentation model. The post-disaster road mask is converted to a road graph (e) using the process described in Section \ref{sec:road_graph}. The difference in mask images (f) is converted to a network graph (g), which is then subtracted from road network graph taken from OpenStreetMap (h), to produce the final post-disaster road network graph (i).}
    \label{fig:block_diagram}
\end{figure*}

\begin{figure*}
    \centering
    \includegraphics[width=\textwidth]{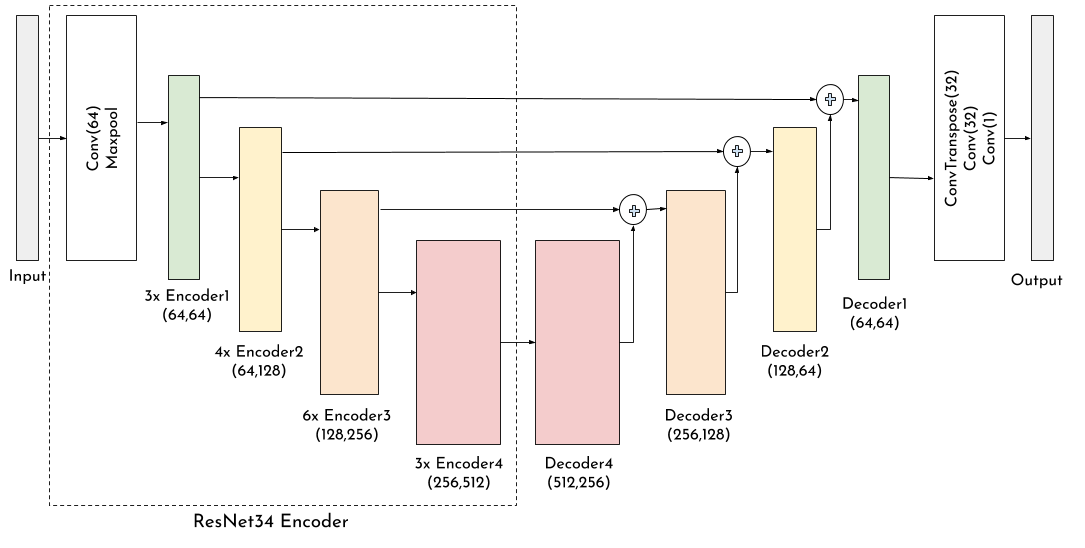}
    \caption{Segmentation model: Input is a satellite image and output is a single channel mask with each pixel representing probability being a road.}
    \label{fig:segmentation_model}
\end{figure*}

\section{METHODOLOGY}
The proposed framework is based on four distinct steps: 1) Road segmentation from satellite images using a CNN; 2) Creation of a road network graph from the segmented images; 3)  Comparing pre-disaster and post-disaster road segments to identify possible changes; 4) Registering the changed segments with the OSM data to get a more realistic road network map. This pipeline is visualised in Fig. \ref{fig:block_diagram} and further described in the following sections.


\subsection{Segmentation} 
\label{sec:segmentation}

\begin{figure}
 \centering
\begin{subfigure}{0.4\textwidth}
    \centering
    \includegraphics[width=\textwidth]{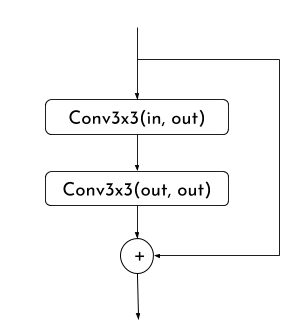}
    \caption{Encoder Block. The number of input and output layers are denoted by 'in' and 'out' respectively.}
    \label{fig:encoder}
\end{subfigure}\hfill
\begin{subfigure}{0.4\textwidth}
    \centering
    \includegraphics[width=\textwidth]{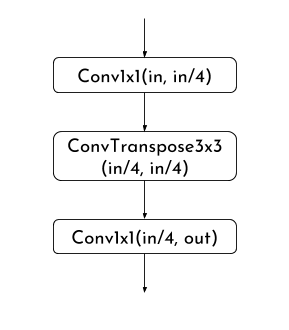}
    \caption{Decoder Block. The number of input and output layers are denoted by 'in' and 'out' respectively.}
    \label{fig:decoder}
\end{subfigure}
\caption{Encoder and Decoder Blocks.}
\end{figure}

We have developed a LinkNet~\cite{Chaurasia2018} based network for the task of semantic road segmentation. It belongs to the family of encoder-decoder segmentation models~\cite{Ronneberger2015} and the architecture is shown in Fig. \ref{fig:segmentation_model}. A ResNet34~\cite{He2016} model pretrained on the ImageNet is used as the encoder since it has been found to yield good performances for the task without excessive computational costs. The encoding part starts with a convolutional block of 64 3x3 filters, followed by a MaxPool layer with a kernel size of 3x3. This is followed by 4 encoding blocks as shown in Fig. \ref{fig:segmentation_model}. Each encoder layer consists of a number of residual blocks as shown in Fig. \ref{fig:encoder}. Each convolutional layer in this case is followed by a batch normalisation and a ReLU layer. The output of each encoder layer feeds into the corresponding decoder layer to help recover the fine details lost in the downsampling by the convolution and pooling layers.

The architecture of a decoder block is shown in Fig. \ref{fig:decoder}, where the transposed convolution layer acts as an upsampling layer. The final layers in the decoding section include a transposed convolution layer, followed by a convolution layer with 32 channels as input and output and a final convolution layer with one output channel corresponding to the class label, either 'road' or 'no road' in this case. The network is trained with a binary cross-entropy loss function using satellite images as input and a probability mask as output, which is compared to the binary mask of the ground truth in order to compute the loss.

\subsection{Generating Road Graphs}
\label{sec:road_graph}
A binary road mask, \(M\), is created from the output of the segmentation network by assuming that any pixels with a probability value greater than 0.5 are road pixels and otherwise are non-road. This mask is then dilated to remove noise and small gaps that may appear during the segmentation process. 

In order for the output to be useful for route extraction, we convert the binary road mask image to a network graph, inspired by the graph theory. Firstly, morphological thinning is used to skeletonise the road mask to obtain a binary mask with a width of one pixel. Since any pixel with more than two neighbours can be assumed to be a node, a road graph is created by traversing the skeleton to identify such nodes. All pixels between two nodes are marked as belonging to an edge. 

\subsection{Comparing Graphs}
\label{sec:graph_comp}
Graph theory offers a number of ways to compare graph similarity. However, these methods are typically based on logical topology whereas in the case of roads, we are also interested in the physical topology. Comparing road network graphs is a non-trivial task since corresponding nodes in the two graphs may have an offset and do not necessarily coincide in spatial coordinates. Furthermore, the edges are not uniform and can have a complex topology.

\begin{figure*}
    \centering
    \includegraphics[width=0.9\textwidth]{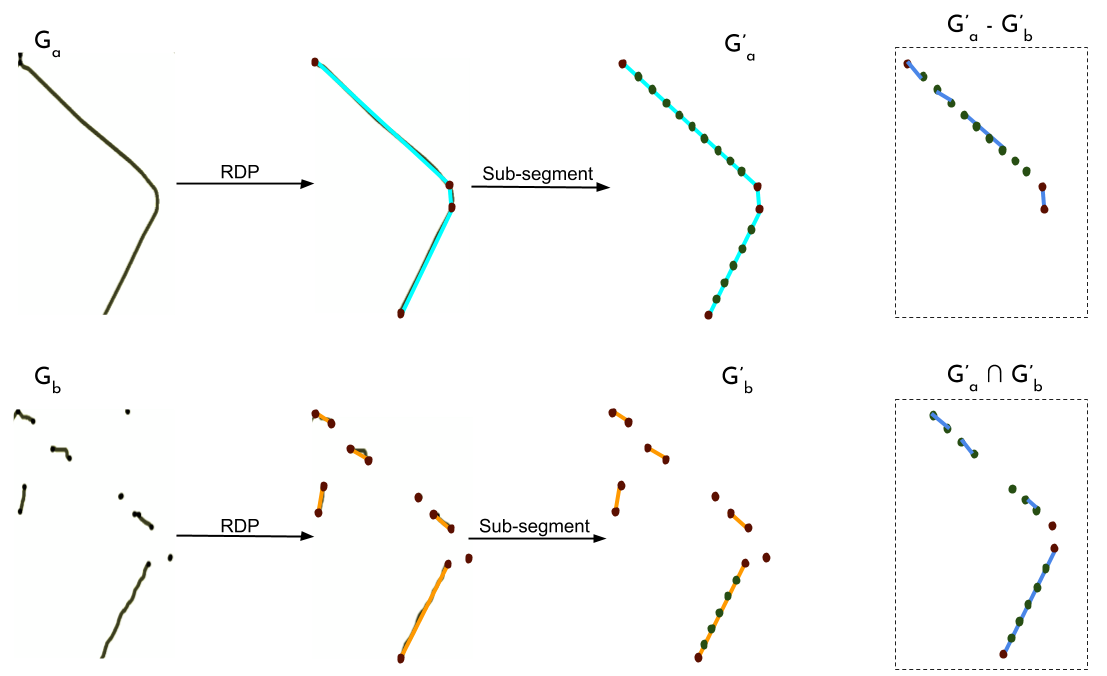}
    \caption{Comparing graphs to find corresponding edges. Initial graphs (\(G_a\) and \(G_b\)) are approximated as combinations of linear segments using RDP~\cite{DOUGLAS1973}. They are further divided into sub-segments of length, \(l\), in graphs \(G'_a\) and \(G'_b\) which can then be compared on a per-segment basis giving the difference between the graphs in the dashed box on the top right and the intersection in the box on the bottom right of the figure.}
    \label{fig:corr}
\end{figure*}

We simplify the graphs, \( G_a\) and \(G_b\), in order to find correspondences as shown in Fig. \ref{fig:corr}. The edges of a graph are approximated with piece-wise linear segments using the Ramer-Douglas-Pecker algorithm~\cite{DOUGLAS1973}. The networks then contain edges that are linear, nodes that are incident to two linear edges and junctions that are incident to three or more edges. The weight of an edge is calculated as the euclidean distance between the vertices of the edge.

Each linear segment is then sliced into smaller sub-segments of a fixed length, \(l\), to create the simplified graphs \( G'_a\) and \(G'_b\), respectively. The sub-segments are compared to find which sub-segments in the two graphs are corresponding as follows:
\begin{equation}
\begin{split}
    \forall e_a, e_b; e_a \in G'_a,  e_b \in G'_b \\
    e_a = \{v_{a1}, v_{a2}\}; e_b = \{v_{b1}, v_{b2}\} \\
    e_a = e_b, \quad \! \textrm{iff} \quad \! |a1 - b1| < l/2 \quad\! \textrm{and} \quad\! |a2 - b2| < l/2
\end{split}
\end{equation}
where the sub-segments in graphs \( G'_a\) and \( G'_b\) are given as \(e_a\) and \(e_b\), which are defined in terms of their two vertices, \( v_{a1}\) and \(v_{a2}\), and \( v_{b1}\) and \(v_{b2}\), respectively. The euclidean distance between two vertices is given by \(|a1 - b1|\) where \(a1\) and \(b1\) are the coordinates of the first vertices of \( v_{a1}\) and \( v_{b1}\) respectively. Essentially, two segments, \(e_a\) and \(e_b\), are assumed to be corresponding if both the vertices of \(e_a\) are within a certain distance of both vertices of \(e_b\). The corresponding segments can be used to find the intersection of the two graphs as shown in Fig. \ref{fig:corr}.

\subsection{Post Disaster Mapping}
In an ideal scenario, the road graphs generated from Section \ref{sec:road_graph} can be used directly for mapping in the post-disaster imagery. However, due to non-ideal segmentation masks, the network graphs are often missing available connections. Herein, we propose to use the difference in the output masks from pre-disaster and post-disaster imagery in conjunction with the OSM data to obtain a more realistic map. 

Both pre-disaster and post-disaster images are used to obtain road masks as described in Section \ref{sec:segmentation}. In order to compensate for some image registration errors, the post-disaster mask \(M_{post}\) is used as a sliding window over the pre-disaster mask, \(M_{pre}\) to find the point where their correlation is the highest. This helps partially correct the alignment of the images.

 Both masks are dilated to deal with small errors during the segmentation and the difference between these maps, \(M_{diff}\) is computed as follows: 
 \begin{equation}
     M_{diff_{p}} = 
     \left\{\begin{matrix}
      1 & \text{if $M_{pre_{p}}$=1 and $M_{post_{p}}$=0} \\
      0 & \text{otherwise}
\end{matrix}\right.
\end{equation}
where \(M_{pre_p}\) is the value of pixel \(p\) in the pre-disaster mask and \(M_{post_p}\) is the value of the corresponding pixel in the post-disaster mask. This function computes the change where a road existed in the pre-disaster mask but is absent in the post-disaster mask.
 
 All output lines in \(M_{diff}\) that are thinner than a certain threshold can be assumed as noise due to registration error and are removed using erosion. Morphological opening is carried out to remove further noise from the image. The final image provides the routes that changed due to the disaster. A road graph \(G_{diff}\) is generated from this image following the process described in Section \ref{sec:road_graph}. An ideal pre-disaster road network \(G_{pre-ideal}\) for the region is obtained from the publicly available OSM dataset. Although the OSM data is not completely accurate~\cite{Mattyus2017}, our experiments have found that using the prior knowledge from the OSM is useful for creating a more robust output.

For each edge in \(G_{diff}\), the closest edge in \(G_{pre-ideal}\) is found as explained in Section \ref{sec:graph_comp} and the cost of the corresponding edge, \(C_e\), is updated according to the following equation:

\begin{equation}
\label{eq:cost}
C_{e} = \alpha \times \frac{s_{e,diff}}{d^2}
\end{equation}
where \(s_{e,diff}\) is the size of the missing segment in \(G_{diff}\), \(d\) is the distance between the missing point in \(G_{diff}\),  and the corresponding edge in \(G_{pre-ideal}\) and \(\alpha\) is an impact factor based on the scale of disaster. The value of \(\alpha\) can be varied from 1, implying no effect from disaster, to \(\infty\) in areas where the roads are completely disconnected or missing such as in areas that might have been washed away by floods.



\section{EXPERIMENTS}
\subsection{Datasets}

DigitalGlobe provides high resolution satellite imagery in wake of natural disasters as part of its Open Data Initiative~\cite{DIGI} to support disaster recovery. We have identified a dataset from DigitalGlobe that has both pre-event and post-event imagery and visible damages to human settlements due to disasters. It is based on the earthquake and tsunami that devastated Sulawesi Island, Indonesia on 28th September, 2018. We extracted an area of approximately \SI{45}{\kilo\metre\squared} around Palu city. An area of \SI{14}{\kilo\metre\squared} with noticeable disaster damage was used for testing. 

\begin{figure*}
    \centering
    \includegraphics[width=0.7\textwidth]{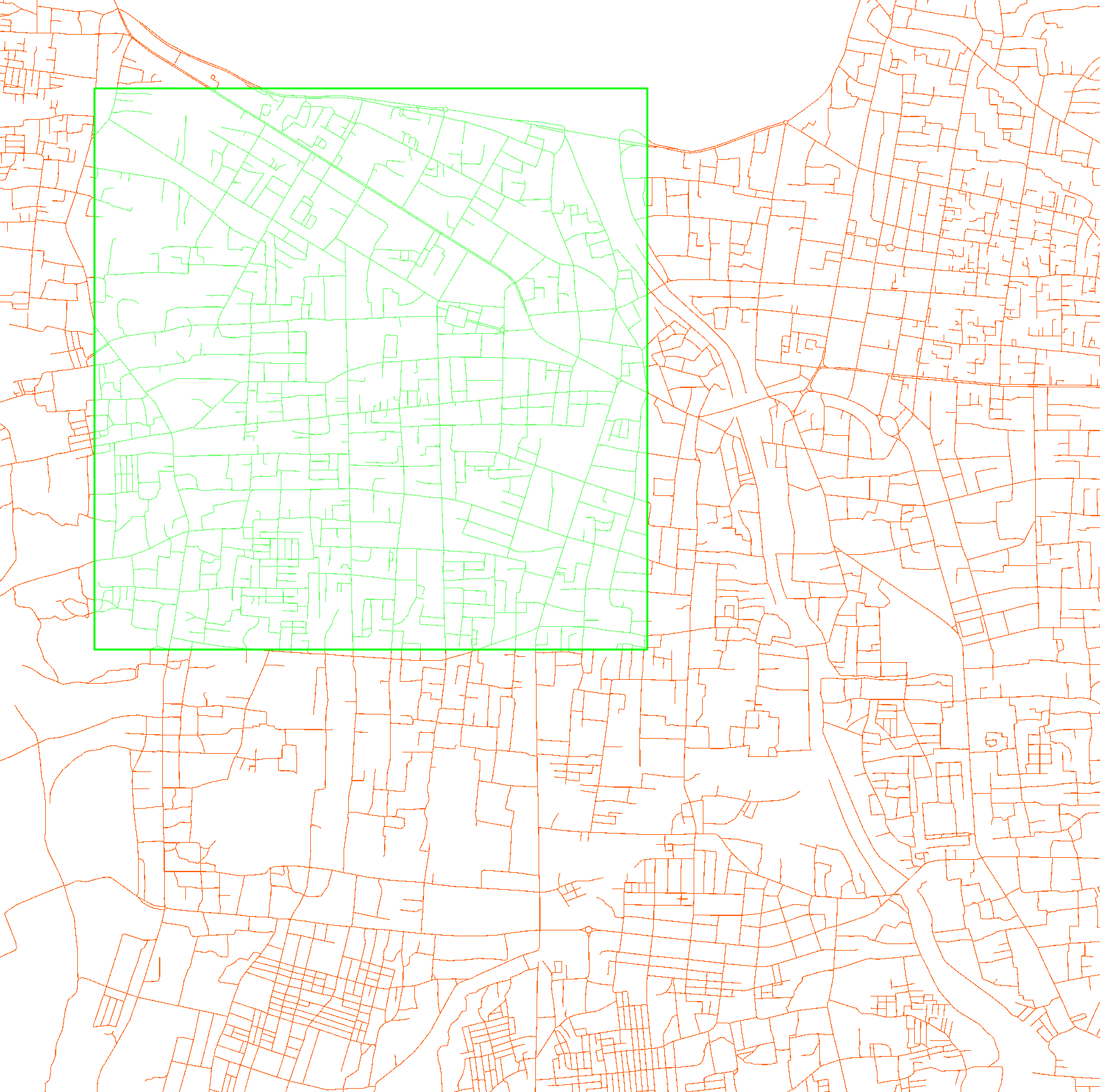}
    \caption{Road map showing the extent of the dataset used. The section outlined in the green box was used as the test dataset and the section in orange was used for training in the case of `\textit{splitDataset}'.}
    \label{fig:dataset}
\end{figure*}

Following the standard practice of using separate areas for training and testing, the first experiment used imagery over an area of \SI{31}{\kilo\metre\squared} for training and validation while the remaining area of \SI{14}{\kilo\metre\squared} for testing. The split is shown in Fig. \ref{fig:dataset}, where the area showing noticeable disaster damage was used as the test area. This dataset has been referred to as `\textit{splitDataset}' in the following sections.

Since the training dataset only used pre-disaster imagery but the inference was done over both the pre-disaster and post-disaster imagery, a second set of experiments used the entire \SI{45}{\kilo\metre\squared} pre-disaster imagery for training. This dataset has been called `\textit{wholeDataset}' in the experimental section. In both cases, a random subset of 10\% of the training data was used  for validation. 




We also downloaded the publicly available data from the OSM for the entire region and extracted all the features marked as roads from this dataset, including all highways, lanes and bicycle paths. The available data from the OSM only includes labels from the pre-disaster imagery in a vector form. 

In order to form suitable data for training the network, we converted the vector road labels to a raster format by converting the lat-long coordinates to pixel coordinates and using a \SI{2}{\metre} buffer around the lines identifying roads for the mask. The roads in the post-disaster imagery were manually labelled for testing purposes.

\subsection{Metrics}
\label{sec:metrics}

There are two primary types of metrics for road extraction: the first is a pixel-wise metric to quantify the performance of the segmentation algorithm, and the second is based on the structure and completeness of the graph.

When defining road networks, a pixel-wise metric such as intersection-over-union (IoU) is not suitable since smalls gaps in the output road mask may cause a small error in IoU but, in reality, can lead to large detours if used in a graph for navigation purposes. Graph-based metrics are more difficult to define and quantify since determining how similar two topological graphs are is a non-trivial problem. 

In this case, we first compare graphs by using their sub-segments as described in Section \ref{sec:graph_comp} and report the standard metrics of precision, recall and F-score defined as follows:

\begin{equation}
\begin{split}
p=\frac{TP}{TP+FP} \\
r=\frac{TP}{TP+FN}  \\
F_{score} = 2\times \frac{p\times r}{p + r}
\end{split}
\end{equation}
where \(TP\) is true positive rate of the segments, \(FN\) false negative rate, \(FP\) false positive rate, and \(F_{score}\) is a measure of the overall accuracy.

These metrics provide a measure of similarity of the road networks. However, they do not take road connectivity into account, which is of particular importance when calculating routes. Hence, we also report the metrics proposed in \cite{Wegner2015} by generating a large set of source-destination pairs and finding the shortest paths for those pairs in the ground truth graph, \(G_{post-ideal}\), and the predicted graph. 

Based on the length of the extracted paths, it is possible to measure if the two graphs are identical as the path lengths should be similar. If the extracted path length is too short compared to the original graph, there will be incorrect shortcuts predicted in the network. Conversely, if the length of the output path is too long or there are no connections, there will be gaps in the graph where there should be roads.

\subsection{Baseline}

We have compared our method to the basic version of DeepRoadMapper (DRM)~\cite{Mattyus2017} that uses a ResNet55 based encoder-decoder network and a soft-jaccard loss for training. The output is similar to the post-disaster mask in our proposed framework shown in Fig. \ref{fig:block_diagram}(e).

\subsection{Training Details}

The satellite images and corresponding masks were clipped to 416x416 pixels. Only the pre-disaster imagery was used as training images with the corresponding OSM information as labels. The post-disaster imagery was used purely for inference. The network was trained using the Adam optimiser~\cite{Kingma2015} with a learning rate of 0.0001 for 100 epochs and a batch size of 12. A pretrained ResNet34 was used to initialise the encoder and the He initialisation~\cite{He2015b} was used for the decoder.


\section{Results}

\begin{figure*}[h!]
\begin{tabular}{|c|c|c|}
\hline
\textbf{Ground Truth} & \textbf{Baseline} & \textbf{Ours} \\ & &\\

\begin{subfigure}[b]{0.31\textwidth}
\includegraphics[width = \textwidth]{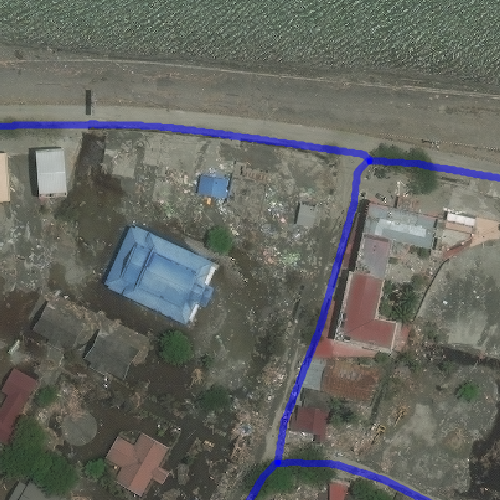} 
\end{subfigure} &
\begin{subfigure}[b]{0.31\textwidth}
\includegraphics[width = \textwidth]{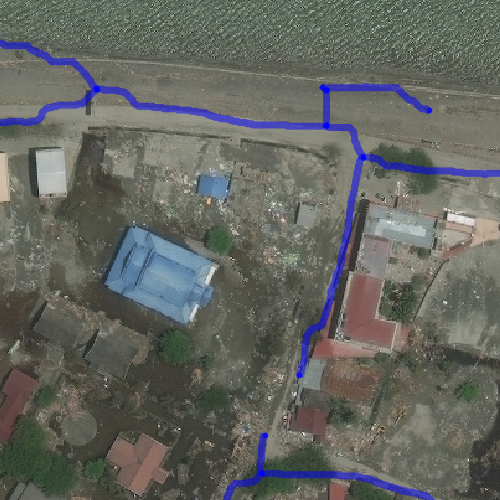} 
\end{subfigure} &
\begin{subfigure}[b]{0.31\textwidth}
\includegraphics[width = \textwidth]{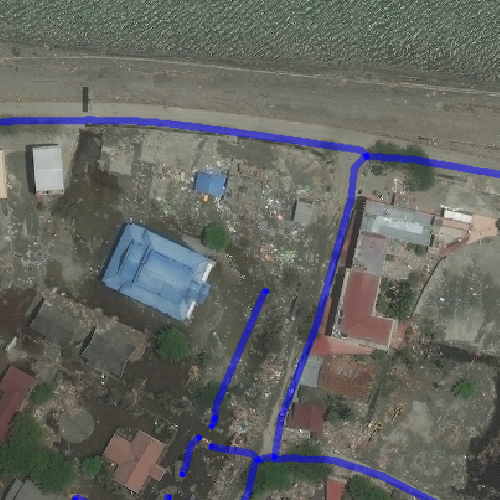} 
\end{subfigure} \\[0.3cm]

\begin{subfigure}[b]{0.31\textwidth}
\includegraphics[width = \textwidth]{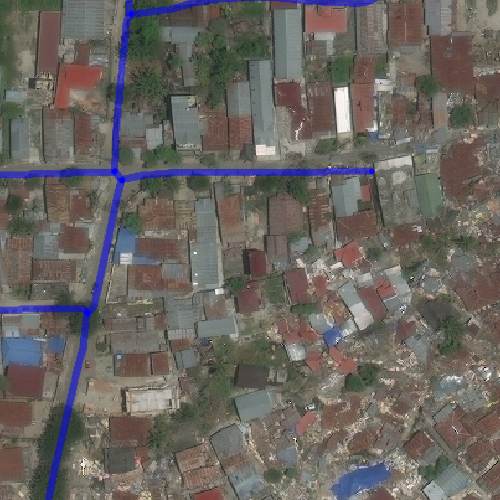} 
\end{subfigure} &
\begin{subfigure}[b]{0.31\textwidth}
\includegraphics[width = \textwidth]{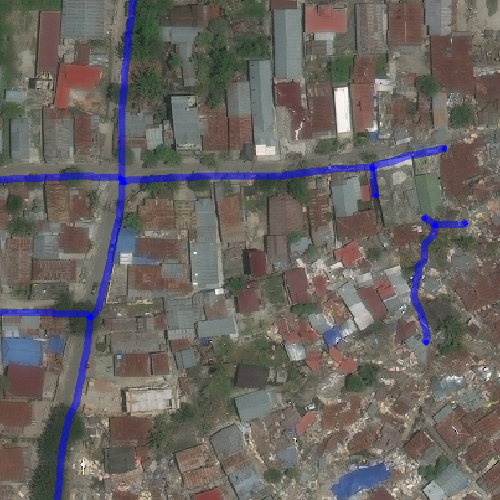} 
\end{subfigure} &
\begin{subfigure}[b]{0.31\textwidth}
\includegraphics[width = \textwidth]{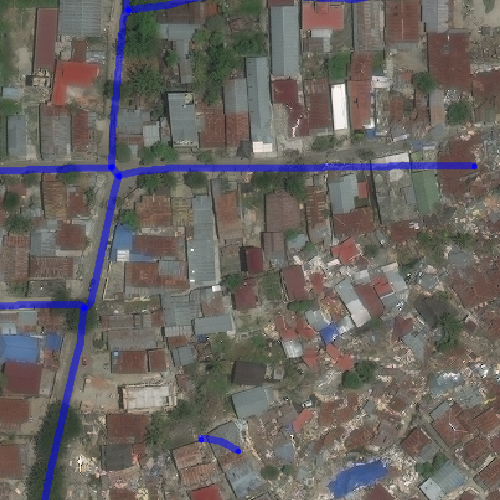} 
\end{subfigure} \\[0.3cm]

\begin{subfigure}[b]{0.31\textwidth}
\includegraphics[width = \textwidth]{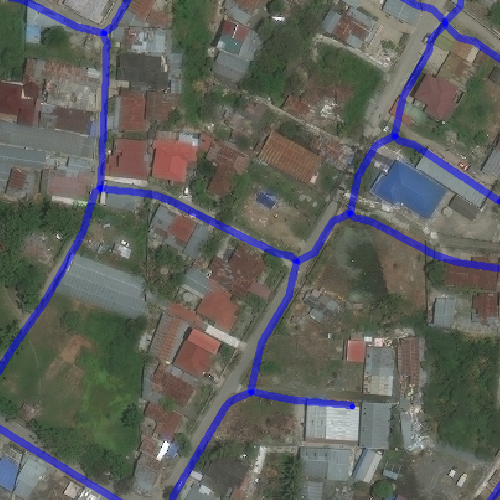} 
\end{subfigure} &
\begin{subfigure}[b]{0.31\textwidth}
\includegraphics[width = \textwidth]{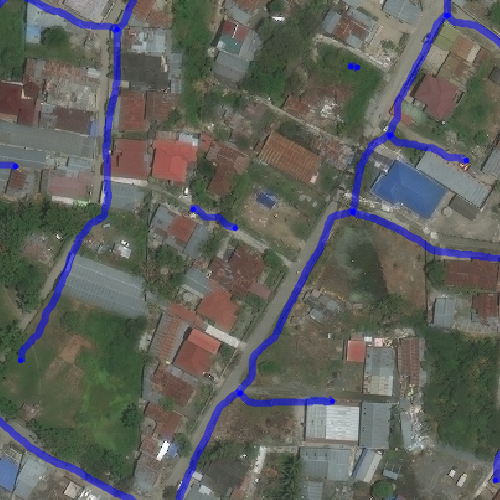} 
\end{subfigure} &
\begin{subfigure}[b]{0.31\textwidth}
\includegraphics[width = \textwidth]{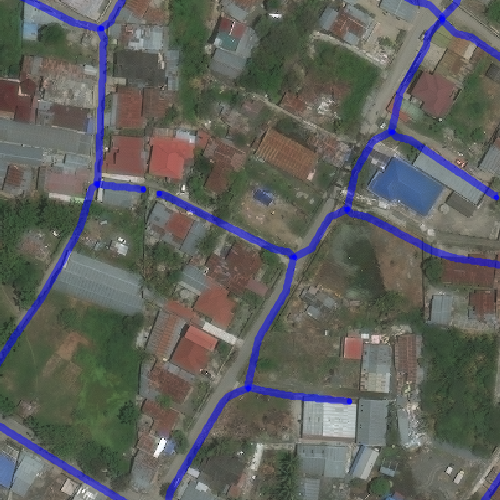} 
\end{subfigure} \\[0.3cm] \hline

\end{tabular}
\caption{Visualisation of results with extracted roads from post-disaster imagery shown in blue. \textit{Left}: GT (Manually Labelled). \textit{Middle}: Baseline DRM~\cite{Mattyus2017}. \textit{Right}: OSM Diff (Ours)}
\label{fig:vis}
\end{figure*}

\subsection{Quantitative Results}

\begin{table}[]
\caption{\textbf{Quality, Completeness, Correctness of Sub-Segments}}
\centering
\begin{tabular}{| l | c| c | c | c | c | c | c |}
\hline
\textbf{Method} & \textbf{ Dataset Split } & \textbf{ TP } &\textbf{ FP } &\textbf{ FN } & \textbf{ Precision } & \textbf{ Recall } & \textbf{ F-score } \\ \hline
DeepRoadMapper~\cite{Mattyus2017} & splitDataset & 5899 & 1011 & 597 & 0.85 & 0.90 & 0.88 \\ \hline 
DeepRoadMapper~\cite{Mattyus2017} & wholeDataset & 6073 & 995 & 423 & 0.86 & 0.93 & 0.89 \\ \hline 
OSM Diff (\textbf{Ours}) & splitDataset & 6453 & 395 & 43 & 0.94 & 0.99 & 0.96 \\ \hline
OSM Diff (\textbf{Ours}) & wholeDataset & 6451 & 387 & 45 & 0.94 & 0.99 & 0.96 \\ \hline
\end{tabular}
\label{tab:prec}
\end{table}

The performance in terms of precision, recall and accuracy is shown in Table \ref{tab:prec}, where the results of DRM~\cite{Mattyus2017} are also compared with that of the proposed approach, termed as \textit{OSM Diff}. In this case, the value of \(\alpha\) was set to \(\infty\) so the changed road segments were completely removed from the road network. This allowed for a fair comparison since all broken roads have been marked as disconnected in the ground truth of the dataset. As can be seen from the results, our method outperformed the baseline by a large margin for both splits of the dataset. This was because that a number of the road segments were missed by the segmentation approach in \cite{Mattyus2017}. Note that our method benefited from the prior knowledge of OSM and had better connectivity than methods that assumed no prior knowledge other than a training dataset.

Another point worth noting is that the test results were similar across the datasets, regardless of the method used. This is possibly because the training was always done on the pre-disaster imagery whereas the test dataset included only the post-disaster imagery. Hence, even though the areas overlapped, the training and testing dataset were disparate and sufficiently large to allow for generalisation.

\begin{table}[]
\caption{\textbf{Connectivity Results}. Correct implies that the shortest paths are similar in length to the ground truth, \textit{No Connections} is where there was no possible path, \textit{Too Short} is where the paths were too short compared to the ground truth and \textit{Too Long} is where the path was too long. All values are given as percentages.}
\centering
\begin{tabular}{| l | l | l | l | l | l |}
\hline
\textbf{Graph Type} & \textbf{ Dataset Split} & \textbf{ Correct } &  \textbf{\begin{tabular}[c]{@{}l@{}} No \\ Connections \end{tabular}} & \textbf{\begin{tabular}[c]{@{}l@{}} Too     \\ Short \end{tabular}} & \textbf{\begin{tabular}[c]{@{}l@{}} Too     \\ Long \end{tabular}} \\ \hline
DeepRoadMapper\cite{Mattyus2017} & splitDataset  & 25.81  & 53.11 & 2.95 & 18.08 \\ \hline
DeepRoadMapper\cite{Mattyus2017} & wholeDataset  & 41.31  & 40.17 & 6.21 & 12.26 \\ \hline
OSM Diff(\textbf{Ours})& splitDataset  & 68.97  & 20.05  & \textbf{1.53} & 9.36 \\ \hline
OSM Diff(\textbf{Ours})& wholeDataset  & 73.38 & 16.76 & 2.03 & 7.72 \\ \hline 
OSM Weighted Diff(\textbf{Ours}) & wholeDataset &\textbf{ 86.59} & \textbf{0} & 8.18 & \textbf{5.13} \\ \hline
\end{tabular}
\label{tab:conn_res}
\end{table}

We report the connectivity results as described in Section \ref{sec:metrics} in Table \ref{tab:conn_res}. Our method outperformed DRM by a large margin. This was again due to a number of missing connections. As can be seen from the table, the path planner was unable to find any paths for about 16\% of the pairs in the output from \textit{OSM Diff} whereas DRM had over twice the number of missing connections. 

The ablation study across different dataset splits for connectivity results showed that both methods performed measurably better when trained with a larger dataset. This was contrary to the precision and recall metrics, which were found similar across the datasets. These results show that the segmentation network performed better at identifying connected segments when provided with more training data.

We also report the results for \textit{OSM Weighted Diff} where the value of \(\alpha\) was set to 5. This allowed for a higher number of correct paths with no missing connections since the network graph was similar to the OSM graph but with a higher cost on affected roads. However, this method gave a larger number of 'too short' paths since it allowed paths which might be impossible to traverse in the post-disaster scenario due to roads that might have been flooded or washed away. Fig. \ref{fig:vis} shows the outputs of the proposed method along with the ground truth and the results from \cite{Mattyus2017}.


\subsection{Qualitative Results}


\begin{figure}[thpb]
      \centering
      \begin{subfigure}[]{.45\textwidth}
      \centering
          \includegraphics[width=0.8\textwidth]{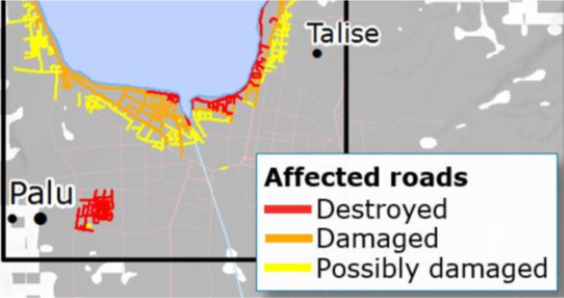}
      \end{subfigure}
    \begin{subfigure}[]{.5\textwidth}
    \centering
    \includegraphics[width=\textwidth]{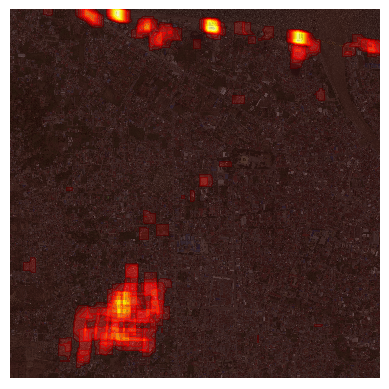}
    \end{subfigure}
      \caption{\textit{Left}: Map of affected roads extracted from \cite{Centre2018}. \textit{Right}: Heatmap of our results indicating the severity of the impact overlaid onto the satellite image. Yellow (Extremely severe) \(>\) Red (Less severe)}
      \label{fig:heatmap}
 \end{figure}

The difference in the segmentation masks, \(M_{diff}\), can be used for identifying the most impacted areas. An area under consideration can be divided into small grids of a fixed size and all pixels in a grid summed to get an estimate of how affected the area is. This was done over the results from our experiments on the Palu imagery and the output has been plotted as a heat map shown in Fig. \ref{fig:heatmap}. 

The results matched the conclusions of the European Commission report~\cite{Centre2018} on the impact of the disaster. The earthquake caused soil liquefaction in the south-west region of Palu, which can be seen as an area of major impact. The coast was also mostly impacted due to the tsunami, which again can be seen in the figure.

\section{CONCLUSIONS}

This paper outlines a framework for identifying road networks in post-disaster scenarios using both satellite imagery and OSM data. It is based on the use of CNNs for road segmentation and graph theory for comparing the changes detected from pre-disaster and post-disaster satellite imagery to leverage knowledge from OSM. This mapping process is currently done manually, while the proposed method can reduce the annotation time down from days to minutes, enabling provision of timely assistance to subsequent relief and rescue work.

The proposed method has been tested on a dataset of Palu, Indonesia from 2018 around the time it was struck by a tsunami and an earthquake. Both quantitative and qualitative results were promising in identifying accessible routes in the region, and the method also successfully identified the highly affected areas in the city. 

The work can be further improved by identifying the reasons for the broken roads and updating the cost function accordingly. For example, standing water or road debris are identified as obstacles and can have a lower cost in the network map but areas that have had landslides should have a much higher cost.

\section*{ACKNOWLEDGMENT}

The authors would like to thank Drs. Andrew West and Thomas Wright for their valuable insights.

{\small
\bibliographystyle{splncs04}
\bibliography{main}
}
\bibliographystyle{splncs04}

\end{document}